\pdfoutput=1

\documentclass[11pt]{article}

\usepackage{EMNLP2023}

\usepackage{times}
\usepackage{latexsym}
\usepackage{booktabs}
\usepackage{graphicx}
\usepackage{amsmath}
\usepackage{amssymb}
\usepackage{hyperref}
\usepackage{multirow} 
\usepackage{siunitx}  
\usepackage{todonotes}  
\usepackage{booktabs, siunitx, adjustbox} 
\usepackage[T1]{fontenc}

\usepackage[utf8]{inputenc}

\usepackage{microtype}

\usepackage{inconsolata}

\title{Context-aware Rotary Position Embedding}

\author{
  Ali Veisi \quad
  Delaram Fartoot \quad
  Hamidreza Amirzadeh \\
  Axiom Lab \\
  \texttt{\{ali.veisi, h.amirzadeh, d.fartoot\}@axiomlab.org}
}

\begin{document}
\maketitle
\begin{abstract}
Positional encoding is a vital component of Transformer architectures, enabling models to incorporate sequence order into self-attention mechanisms. Rotary Positional Embeddings (RoPE) have become a widely adopted solution due to their compatibility with relative position encoding and computational efficiency. However, RoPE relies on static, input-independent sinusoidal frequency patterns, limiting its ability to model context-sensitive relationships. In this work, we propose CARoPE (Context-Aware Rotary Positional Embedding), a novel generalization of RoPE that dynamically generates head-specific frequency patterns conditioned on token embeddings. This design introduces token- and context-sensitive positional representations while preserving RoPE’s efficiency and architectural simplicity. CARoPE computes input-dependent phase shifts using a bounded transformation of token embeddings and integrates them into the rotary mechanism across attention heads. We evaluate CARoPE on the FineWeb-Edu-10B dataset using GPT-2 variants trained on next-token prediction tasks. Experimental results show that CARoPE consistently outperforms RoPE and other common positional encoding baselines, achieving significantly lower perplexity, even at longer context lengths. Additionally, CARoPE enables faster training throughput without sacrificing model stability. These findings demonstrate that CARoPE offers a scalable, expressive, and efficient upgrade to existing positional encoding strategies in Transformer models.
\end{abstract}

\section{Introduction}
\label{intro}

Transformer architectures have revolutionized the field of deep learning~\cite{vaswani2017attention}, achieving state-of-the-art performance across a wide range of tasks in natural language processing~\cite{devlin-etal-2019-bert,liu2019roberta,chowdhery2023palm,team2023gemini,touvron2023llama,achiam2023gpt}. A key component of their success is the self-attention mechanism, which enables the model to dynamically capture relationships between elements in a sequence, regardless of their distance. However, unlike traditional sequence models such as Recurrent Neural Networks (RNNs) ~\cite{sherstinsky2020fundamentals} or Convolutional Neural Networks (CNNs)~\cite{gehring2017convolutional}, transformers lack an inherent sense of order or position~\cite{yun2019transformers}. This makes \textit{positional encoding} a crucial component, as it injects position-related information into the model to enable sequence-aware processing.

Over the years, several strategies for positional encoding have been proposed. These include fixed sinusoidal embeddings~\cite{vaswani2017attention}, learnable absolute position embeddings~\cite{devlin-etal-2019-bert}, relative position encodings~\cite{press2021train,raffel2020exploring}, and rotary positional embeddings (RoPE)~\cite{su2024roformer}. Among these, RoPE has become one of the most widely adopted approaches due to its compatibility with self-attention and ability to encode relative positions through rotation-based transformations.

RoPE works by rotating the query and key vectors within the multi-head attention mechanism using fixed sinusoidal frequencies. Although effective, RoPE still relies on predefined static frequency patterns that are uniform across different inputs and attention heads. As a result, it remains position-dependent but not token- or context-dependent, limiting its expressiveness in modeling more nuanced sequence structures.

In this work, we propose \textbf{CARoPE} (\textit{Context-Aware Rotary Positional Embedding}), a novel enhancement of RoPE that introduces \textit{dynamic, input-dependent frequency values} for each attention head. By making frequency generation sensitive to the input content, CARoPE enables the model to adaptively encode positional information in a way that reflects both the position and the underlying context. This results in more expressive and flexible positional representations that are conditioned on the input context and vary across attention heads.

Unlike RoPE's fixed sinusoidal formulation, CARoPE learns a nonlinear transformation of the input embeddings to generate head-specific frequency patterns, which are then integrated into the rotary positional mechanism. This context-aware extension enables richer, token-sensitive position encoding without sacrificing the efficiency and compatibility of the original RoPE framework. We assess the effectiveness of our approach across multiple benchmark datasets, employing GPT-2 variants for the standard next-token prediction task. CARoPE consistently outperforms existing positional encoding methods, including RoPE, and achieves lower perplexity in generated sequences.



\section{Proposed Method}

We formulate CARoPE as a generalization of Rotary Positional Embedding (RoPE), designed to introduce context-dependent positional modulation within the attention mechanism. While RoPE encodes relative position through fixed sinusoidal rotations, CARoPE replaces these static frequencies with dynamic, token- and head-specific alternatives.

To motivate our method, we first reinterpret standard RoPE through the lens of phase accumulation. In RoPE, the position-dependent rotation applied to each embedding pair is defined as:
\[
\phi_i(m) = m \cdot \theta_i,
\]
where $m$ is the sequence position and $\theta_i = 10000^{-2i/d}$ is the fixed frequency assigned to the $i$-th embedding pair in a $d$-dimensional space. This can be reformulated as a cumulative sum:
\[
\phi_i(m) = \sum_{t=1}^m \theta_i.
\]
Noting that $\theta_i$ follows a geometric progression, the phase term becomes:
\[
\phi_i(m) = \sum_{t=1}^m \theta_1^i = m \cdot \theta_1^i,
\]
revealing that each rotational component increases exponentially with dimension.

CARoPE generalizes this formulation by replacing the fixed base frequency $\theta_1$ with a learned, input-dependent function $f(x_t)$, where $x_t \in \mathbb{R}^d$ is the embedding of the token at position $t$. The generalized phase term becomes:
\[
\phi_i^{(h)}(m) = \sum_{t=1}^m f(x_t)_h^i,
\]
where $h$ indexes the attention head, and $f(x_t)_h \in (0, 1)$ is a learned, bounded scalar frequency specific to head $h$ and token $x_t$. This formulation maintains the exponential dimension-wise progression of RoPE but allows the frequency to vary across both tokens and heads, yielding context-aware phase accumulation.

The frequency modulation function $f$ is implemented as:
\[
f(x_t) = \frac{1}{\text{softplus}(x_t W) + 1},
\]
where $W \in \mathbb{R}^{d \times h}$ projects the token embedding to $h$ scalar values, one per head. The softplus activation ensures positivity, while the inverse squashing maps outputs to the interval $(0, 1)$, promoting stability when raised to higher powers.

After computing $\phi_i^{(h)}(m)$ for each position, head, and dimension, we construct sinusoidal components:
\[
\cos\big(\phi_i^{(h)}(m)\big), \quad \sin\big(\phi_i^{(h)}(m)\big),
\]
which are then applied to the query and key vectors using the standard RoPE formulation.

To preserve stability and enable efficient training, we initialize CARoPE using the standard RoPE formulation. Since RoPE corresponds to a special case of CARoPE. This initialization ensures the model begins with a valid and expressive positional prior.

\section{Experiment Setup}
\subsection{Datasets}
For training, we use the FineWeb dataset \cite{penedo2024fineweb}, a large-scale dataset (15 trillion tokens) for LLM pretraining, derived from 96 CommonCrawl snapshots. FineWeb has been shown to produce better-performing LLMs than other open pretraining datasets \cite{penedo2024fineweb}. More specifically, we use a 10B sample of the FineWeb-Edu dataset, which consists of 1.3T tokens from educational web pages filtered from the FineWeb dataset. We allocate 9.9B tokens for training and 0.1B for evaluation. 
For evaluation, we use the test set of FineWeb-Edu.

\begin{table*}
    \centering
    \footnotesize
    \setlength{\tabcolsep}{6pt} 
    \begin{tabular}{lcccc}
        \toprule
        \multicolumn{5}{c}{\textbf{GPT-Small models}} \\
        \midrule
        Sequence Length & RoPE & CARoPE & Learnable & Sinusoidal \\
        \midrule
        512   & 21.31 & \textbf{21.23} & 21.90 & 22.14 \\
        1024  & 56.61 & \textbf{21.39} & - & 166.18 \\
        \bottomrule
        \midrule
        \multicolumn{5}{c}{\textbf{GPT-Tiny models}} \\
        \midrule
        Sequence Length & RoPE & CARoPE & Learnable & Sinusoidal \\
        \midrule
        512   & 29.33 & \textbf{28.99} & 30.48 & 30.62 \\
        1024  & 81.27 & \textbf{36.74} & - & 223.28 \\
        \bottomrule
    \end{tabular}
\caption{
Perplexity comparison on the FineWeb-Edu-10B evaluation set. The first row reports results from GPT-Small models, and the second row shows results from GPT-Tiny models. All models were trained for 19k steps on the FineWeb-Edu-10B training set with a context length of 512.
}
\label{tab:main_results}
\end{table*}

\subsection{Settings}
For all next-token prediction tasks, we use the GPT-2 variants \cite{brown2020language}. For the FineWeb-Edu-10B dataset, we use its small version (12 layers, 10 heads, and a hidden dimension of 768) with 124M parameters, 
and a tiny version of GPT-2 (44M parameters) with 6 layers, 8 heads, and a hidden dimension of 512. The evaluation metric is perplexity (PPL), and we train the models with sequence length of 512. All the models are trained on two H100 GPUs with 80G GPU RAM. Training settings are the same as those used for GPT-2 \cite{radford2019language}. Gradients are updated after processing 524,288 tokens and vocab size is 50304. For training on the FineWeb-Edu-10B dataset, we run 19k steps (\textasciitilde{}1 epoch) with batch sizes of 64, and 32 for the tiny, and small models, respectively. The learning rate starts at 0.0006, with a linear warmup over 750 steps, followed by cosine decay to a minimum of 0.00006.

\subsection{Baselines}
We compare our method against the following positional encoding approaches:

\textbf{Learnable} \cite{vaswani2017attention}: A trainable additive positional encoding (APE) where each position is associated with a learned embedding. The number of positions is fixed and predefined during training.

\textbf{Sinusoidal} \cite{vaswani2017attention}: A fixed APE used in early Transformer models \cite{vaswani2017attention, baevski2018adaptive, ott2018scaling, lewis2021base}.

\textbf{RoPE} \cite{su2024roformer}: A non-learnable relative positional encoding (RPE) widely adopted in LLMs such as GPT-2 \cite{brown2020language}, LLaMA \cite{touvron2023llama}, PaLM \cite{chowdhery2023palm}, and Gemma \cite{team2024gemma, team2024gemma2}.

\section{Results}
\label{sec:results}

Table~\ref{tab:main_results} reports the perplexity of models trained with sequence length of 512 and different positional encoding strategies on the FineWeb-Edu-10B evaluation set. Across both GPT-Small and GPT-Tiny variants, CARoPE consistently outperforms RoPE, achieving notably lower perplexity, especially at longer sequence lengths. For example, at a sequence length of 1024, CARoPE reduces perplexity by more than 60\% compared to RoPE in the GPT-Tiny model (36.74 vs. 81.27). This demonstrates CARoPE's ability to generalize better over longer contexts.

The results validate the effectiveness of dynamic, input-dependent frequency modulation in enhancing positional representation. Notably, CARoPE not only achieves better perplexity but also enables faster training, processing approximately 0.76 million tokens per second compared to 0.63 million for RoPE in GPT-Small models.

\section{Conclusion}

We presented CARoPE, a context-aware extension of Rotary Positional Embeddings that introduces input- and head-dependent frequency modulation. By dynamically adapting to token content, CARoPE improves the expressiveness of positional encoding with minimal overhead. Our experiments demonstrate consistent gains over RoPE across model sizes and sequence lengths, highlighting its effectiveness for enhancing Transformer-based language models.


\bibliography{main}
\bibliographystyle{acl_natbib}

\newpage
\appendix

\end{document}